\documentclass[10pt]{article}
\usepackage[preprint]{tmlr}

\usepackage{hyperref}
\usepackage{url}
\usepackage{booktabs}
\usepackage{longtable}
\usepackage{array}
\usepackage{calc}
\usepackage{amsmath}
\usepackage{amssymb}
\usepackage{textcomp}
\usepackage{etoolbox}
\usepackage{caption}
\newcounter{none}

\makeatletter
\patchcmd\longtable{\par}{\if@noskipsec\mbox{}\fi\par}{}{}
\makeatother

\AtBeginEnvironment{longtable}{\footnotesize}
\title{The Objective Is the Bottleneck: Latent World Models Encode What Their Planners Cannot Use}

\author{\name Joyjeet Singh \email joyjeetsingh1@gmail.com \\
      \addr Independent researcher \\ ORCID 0009-0005-1512-7439}

\begin{document}
\maketitle

\begin{abstract}
Latent world models are judged by how well they predict, so when planning over
one fails at long horizons the natural reading is that the predictor degrades.
On a reproduction of LeWorldModel on TwoRoom we show the binding constraint is
the planner's objective instead.

The predictor is not the limit: rolled out on real validation clips, its
imagined state seventy-five environment steps ahead is still only 0.189 as
wrong as assuming the world froze, while the planner never imagines beyond
twenty-five. The objective is the limit. Cross-entropy-method planning
minimises squared latent distance, which tracks true distance at r = 0.426,
stops rising by about eighty arena units, and \textbf{decreases} beyond about a
hundred and twenty --- so moving away from the goal can lower the planner's
cost. The information is present throughout: a ridge probe recovers position
from the same frozen embedding at $R^{2}$ 0.9922.

The pathology belongs to the method rather than to one reimplementation. It is
present in the authors' released weights, and across four checkpoints
long-horizon success rank-orders exactly with the quality of this metric and
inversely with one-step prediction accuracy --- a mechanism for a dissociation
the reproduction reports and leaves unexplained.

Replacing only the objective, with nothing retrained and no GPU, lifts goals
reached at offset 100 from 26.0\% to \textbf{98.0\%}, equals the 98.0\% reached at
offset 25, and still reaches 92.0\% under a third of the budget: planning stops
depending on the horizon. The best cost is not the most accurate one. A head
learned from frame separation alone predicts \emph{spatial} distance worse than a
position probe (r = 0.819 against 0.9897) yet plans better, because it charges
24\% more to cross the environment's dividing wall where squared latent
distance charges 4\% \emph{less}. It has learned reachability, not proximity.

We also report where the repair fails: on the authors' weights the learned
cost is beaten by the linear one, 34.0\% against 70.0\%, because a head fit on
encoded frames is evaluated on imagined ones and their predictor drifts far
enough for an MLP to extrapolate. A learned planning cost must be trained on
the distribution the planner scores.

\begin{center}\rule{0.5\linewidth}{0.5pt}\end{center}
\end{abstract}

\section{1 Introduction}\label{introduction}

Latent world models are trained to predict. A model that predicts the next
latent state accurately is taken to have learned the dynamics, and planning
over it is expected to follow. When it does not, the usual suspects are the
predictor's horizon, its capacity, or the amount of training.

This paper reports a case where none of those is the cause. Working from a
published reproduction of LeWorldModel on the TwoRoom environment
(arXiv:2608.10145), where planning reaches 94.0\% of goals twenty-five steps
away and 26.0\% of goals a hundred steps away, we asked what the model would
need in order to close that gap. The answer turned out to be nothing.

The reproduction paper reports the gap and, deliberately, offers no mechanism
for it. It also reports something stranger: across three checkpoints, one-step
prediction accuracy orders short-horizon planning success monotonically and
fails to order long-horizon success at all, with the least accurate checkpoint
the strongest long-horizon planner. Both observations have the same
explanation, and it is not about prediction.

\subsection{Contributions}\label{contributions}

\begin{enumerate}
\def\labelenumi{\arabic{enumi}.}
\item
  \textbf{The predictor is not the bottleneck.} Imagination stays informative to
  at least seventy-five environment steps; the planner uses twenty-five (\S{}3).
\item
  \textbf{The planning objective saturates and then inverts.} Squared latent
  distance --- the quantity CEM minimises --- is not monotone in true distance
  beyond about a hundred and twenty units, so moving away from the goal can
  lower the planner's cost (\S{}4).
\item
  \textbf{The pathology belongs to the method, not to one reimplementation.} It is
  present in the original authors' released weights (\S{}5).
\item
  \textbf{A mechanism for the accuracy/planning dissociation.} Across four
  checkpoints, long-horizon success rank-orders exactly with metric quality
  and inversely with prediction accuracy (\S{}5).
\item
  \textbf{The information is present; only the metric fails.} Position decodes
  from the frozen embedding at $R^{2}$ 0.9922 (\S{}6).
\item
  \textbf{A repair with no retraining}, effective on the authors' own weights, and
  not dependent on the environment having a readable state (\S{}7, \S{}8).
\item
  \textbf{Reachability, not proximity, is the right objective.} A cost that
  predicts spatial distance worse plans better, because it charges for the
  wall. Planning success across three objectives orders by how well each
  captures reachability (\S{}7.3, \S{}8).
\item
  \textbf{A learned cost must be trained on what the planner scores} --- imagined
  embeddings, not encoded frames. Invisible on the checkpoint the method was
  developed on; decisive on the other (\S{}7.4).
\item
  \textbf{A conditional prescription, including where the repair loses.} The
  learned cost wins where the predictor supports it and loses to a linear one
  where it does not (\S{}8.2).
\end{enumerate}

\subsection{Scope}\label{scope}

One environment, one seed per checkpoint, four checkpoints. TwoRoom is a
diagnostic environment and the strongest claims here are about it. \S{}9 states
what that does and does not license.

\section{2 Setup}\label{setup}

All experiments use the checkpoints and evaluation harness released with the
reproduction. The world model is an 18.03M-parameter joint-embedding
predictive architecture: a ViT-Tiny encoder at 224px producing 192-dimensional
embeddings, and a six-layer predictor conditioned on an action encoding.

Planning is cross-entropy-method search over the learned dynamics, with the
released configuration: 300 candidate sequences, 30 refinement iterations, 30
elite, horizon 5, and a receding horizon of 5. The planner emits a
two-dimensional action which the environment executes for \texttt{frameskip} = 5
steps, so a planning horizon of 5 corresponds to 25 environment steps of
imagination.

\textbf{The objective.} CEM scores each candidate by

\[\text{cost} = \lVert \hat{z}_T - z_{\text{goal}} \rVert_2^2\]

the squared Euclidean distance between the imagined final embedding and the
embedding of the goal image. This is the quantity the paper is about.

\textbf{Protocols.} The released material publishes two evaluation protocols. The
repository configuration places the goal 25 frames ahead with a 50-step
budget; the paper's appendix places it 100 frames ahead with a 150-step
budget. We report both, and use the published runs as baselines throughout.
Every comparison below draws the same episodes as its baseline, at the same
seed, which we verified by replicating the draw.

\textbf{Compute.} Every experiment in this paper ran on a four-core laptop CPU. No
model was trained or fine-tuned.

\section{3 The predictor is not the bottleneck}\label{the-predictor-is-not-the-bottleneck}

If long-horizon planning fails because imagination degrades, the degradation
should be visible in a rollout. We rolled the released checkpoint forward
autoregressively on real validation clips under their recorded actions,
comparing each imagined latent against the encoder's true latent for that
frame, and against a static baseline that predicts nothing moves.

{\def\LTcaptype{none} % do not increment counter
\begin{longtable}[]{@{}lll@{}}
\toprule\noalign{}
horizon (planner steps) & environment steps & imagined err / static \\
\midrule\noalign{}
\endhead
\bottomrule\noalign{}
\endlastfoot
1 & 5 & 0.066 \\
5 & 25 & 0.090 \\
10 & 50 & 0.152 \\
15 & 75 & 0.189 \\
\end{longtable}
}

\emph{From \texttt{followup/rollout\_h15\_phase2\_recal.txt}.}

At fifteen planner steps the imagined state is still only 19\% as wrong as
assuming the world froze, and the error grows slowly and smoothly. There is no
collapse anywhere in the range a planner would need.

Horizon 20 could not be measured at all. A clip spanning three history frames
and twenty predictor steps at frameskip 5 needs 102 raw frames, and the
dataset's episodes are capped at 101 --- the same ceiling that makes the
offset-100 protocol degenerate in the reproduction's \S{}4.2.

\textbf{The planner never asks for this.} With horizon 5 it imagines 25
environment steps, toward goals 100 steps away. The model can see three times
farther than it is ever asked to.

\section{4 The objective saturates, then inverts}\label{the-objective-saturates-then-inverts}

The planner's objective is only sensible if latent distance rises with true
distance across the range it must cover. We measured that directly, with no
planning involved: render real recorded positions, encode them, and compare
pairwise latent distance against pairwise true distance.

{\def\LTcaptype{none} % do not increment counter
\begin{longtable}[]{@{}
  >{\raggedright\arraybackslash}p{(\linewidth - 6\tabcolsep) * \real{0.2500}}
  >{\raggedright\arraybackslash}p{(\linewidth - 6\tabcolsep) * \real{0.2500}}
  >{\raggedright\arraybackslash}p{(\linewidth - 6\tabcolsep) * \real{0.2500}}
  >{\raggedright\arraybackslash}p{(\linewidth - 6\tabcolsep) * \real{0.2500}}@{}}
\toprule\noalign{}
\begin{minipage}[b]{\linewidth}\raggedright
true distance (arena units)
\end{minipage} & \begin{minipage}[b]{\linewidth}\raggedright
pairs
\end{minipage} & \begin{minipage}[b]{\linewidth}\raggedright
mean latent distance
\end{minipage} & \begin{minipage}[b]{\linewidth}\raggedright
r within band
\end{minipage} \\
\midrule\noalign{}
\endhead
\bottomrule\noalign{}
\endlastfoot
0--20 & 648 & 15.92 & 0.668 \\
20--40 & 1288 & 18.52 & 0.182 \\
40--60 & 1208 & 19.68 & 0.115 \\
60--80 & 1329 & 20.04 & 0.104 \\
80--100 & 960 & 20.61 & 0.122 \\
100--120 & 733 & 20.68 & $-$0.029 \\
120--150 & 691 & 20.60 & falling \\
150--300 & 283 & 19.93 & falling \\
\end{longtable}
}

\emph{From \texttt{followup/latent\_metric\_phase2\_recal.txt}, 7,140 pairs.}

Overall Pearson r = 0.426. Three things follow.

\textbf{The objective is informative only at short range.} Within-band correlation
is 0.668 under twenty units and below 0.13 beyond forty.

\textbf{It saturates.} Across 80--300 units the mean varies by $-$3.3\%. Two states,
one eighty units from the goal and one three hundred, receive almost the same
score.

\textbf{It inverts.} Past about a hundred and twenty units the relationship turns
negative. A planner minimising this cost can be led \emph{away} from its goal and
score itself as improving.

\subsection{4.1 This is the observed failure}\label{this-is-the-observed-failure}

The reproduction reports that offset-100 failures overshoot. Re-analysing its
committed episode data: 26 of 37 failures finish farther from the goal than
they began, at a median 1.41$\times$ the original separation. Success falls off with
distance exactly where the metric goes blind --- 100\% under 60 units, 37\%
between 60 and 100, and 6\% between 100 and 140.

A saturating, eventually inverting objective predicts all of this: short goals
are solved, long goals are not, and the planner does not merely stall but
travels confidently in the wrong direction.

\section{5 The pathology is the method's, and it explains the dissociation}\label{the-pathology-is-the-methods-and-it-explains-the-dissociation}

\subsection{5.1 The authors' own weights}\label{the-authors-own-weights}

The same measurement on the original authors' released checkpoint --- weights we
did not train:

{\def\LTcaptype{none} % do not increment counter
\begin{longtable}[]{@{}lllll@{}}
\toprule\noalign{}
& Pearson & Spearman & monotone & 80--300 spread \\
\midrule\noalign{}
\endhead
\bottomrule\noalign{}
\endlastfoot
authors' released & 0.388 & 0.423 & no & $-$2.2\% \\
\end{longtable}
}

\emph{From \texttt{followup/latent\_metric\_authors.txt}.}

This is our checkpoint's pathology almost exactly. Whatever produces it is a
property of the method as released, not of a reimplementation choice.

\subsection{5.2 Metric quality orders long-horizon planning}\label{metric-quality-orders-long-horizon-planning}

Measuring all four available checkpoints and placing them beside their
published planning results:

{\def\LTcaptype{none} % do not increment counter
\begin{longtable}[]{@{}
  >{\raggedright\arraybackslash}p{(\linewidth - 10\tabcolsep) * \real{0.1667}}
  >{\raggedright\arraybackslash}p{(\linewidth - 10\tabcolsep) * \real{0.1667}}
  >{\raggedright\arraybackslash}p{(\linewidth - 10\tabcolsep) * \real{0.1667}}
  >{\raggedright\arraybackslash}p{(\linewidth - 10\tabcolsep) * \real{0.1667}}
  >{\raggedright\arraybackslash}p{(\linewidth - 10\tabcolsep) * \real{0.1667}}
  >{\raggedright\arraybackslash}p{(\linewidth - 10\tabcolsep) * \real{0.1667}}@{}}
\toprule\noalign{}
\begin{minipage}[b]{\linewidth}\raggedright
checkpoint
\end{minipage} & \begin{minipage}[b]{\linewidth}\raggedright
one-step error
\end{minipage} & \begin{minipage}[b]{\linewidth}\raggedright
offset 100, budget 150
\end{minipage} & \begin{minipage}[b]{\linewidth}\raggedright
Spearman
\end{minipage} & \begin{minipage}[b]{\linewidth}\raggedright
monotone
\end{minipage} & \begin{minipage}[b]{\linewidth}\raggedright
80--300 spread
\end{minipage} \\
\midrule\noalign{}
\endhead
\bottomrule\noalign{}
\endlastfoot
Run 2 recal & 0.829 & 40/50 = 80.0\% & 0.872 & yes & +38.6\% \\
Run 0 recal & --- & 33/50 = 66.0\% & 0.846 & yes & +34.9\% \\
phase2 recal & 0.116 & 13/50 = 26.0\% & 0.445 & no & $-$3.3\% \\
authors' released & 0.410 & 7/50 = 14.0\% & 0.423 & no & $-$2.2\% \\
\end{longtable}
}

Long-horizon planning success rank-orders \textbf{exactly} with metric quality, and
inversely with one-step prediction accuracy.

This supplies a mechanism for the dissociation the reproduction reports and
declines to explain. The most accurate predictor is the worst long-horizon
planner not because accuracy is irrelevant to planning, but because in these
runs accuracy and metric usability move in opposite directions. Selecting on
prediction loss selects against the geometry the planner depends on.

\subsection{5.3 A correlate, not a cause}\label{a-correlate-not-a-cause}

The split tracks the training pipeline. Both checkpoints with usable geometry
used \texttt{history\_size} 1 and \texttt{action\_dim} 2; both pathological ones --- ours and
the authors' --- use the reference pipeline with \texttt{history\_size} 3 and dense
ten-wide actions.

Effective rank moves with it. The reproduction's committed probe records 18.6
of 192 effective dimensions for Run 0 and 67.8 for phase2, a 3.6$\times$ spread
across the same interval over which the metric degrades. This is a partial
account of dimensions the reproduction explicitly declines to characterise:
whatever they encode, it is not distance-relevant, and it dilutes an L2 taken
over all 192 --- which is consistent with a two-dimensional linear read-out
recovering position at $R^{2}$ 0.99 from the same vector that L2 cannot order.

\textbf{We cannot separate these factors.} History size, action width and pixel
normalisation change together across the four checkpoints, and each
checkpoint is a single seed. This is a correspondence worth explaining, not a
demonstrated cause.

\section{6 The information is present}\label{the-information-is-present}

If the embedding did not contain the geometry, no objective could recover it.
It does. A ridge probe fit on frozen embeddings of 350 rendered positions and
evaluated on 150 held out:

{\def\LTcaptype{none} % do not increment counter
\begin{longtable}[]{@{}lll@{}}
\toprule\noalign{}
& latent L2 & decoded position \\
\midrule\noalign{}
\endhead
\bottomrule\noalign{}
\endlastfoot
Pearson r vs true distance & 0.437 & \textbf{0.9897} \\
monotone in true distance & no & \textbf{yes} \\
spread across 80--300 units & +1.2\% & \textbf{+73.4\%} \\
\end{longtable}
}

with probe $R^{2}$ \textbf{0.9922} held out and mean absolute error \textbf{1.72} arena
units. \emph{From \texttt{followup/probe\_metric\_phase2\_recal.txt}.}

The same holds on the authors' checkpoint: probe $R^{2}$ 0.9627, decoded-position
distance monotone at r = 0.9573 where their latent L2 is neither.

The encoder represents what the planner needs. The objective discards it.

\section{7 Repair}\label{repair}

We change the planner's objective and nothing else. The encoder and predictor
are the released weights, frozen.

\textbf{Decoded-position cost.} Fit a ridge probe from embeddings to position, and
score candidates by distance between decoded positions rather than between
embeddings. The probe is fit at startup on rendered positions.

\textbf{Learned temporal-distance cost.} The decoded-position cost leans on
something TwoRoom happens to offer --- a low-dimensional state a linear probe
can read. To test whether the repair depends on that, we learn a cost from the
only signal a deployed system always has: how many steps apart two observed
frames were.

Training pairs are real recorded frames from the same episode, supervised by
their frame separation. A small MLP maps (z\_t, z\_goal) to predicted
steps-to-reach, symmetrised over both orderings. Episodes are split disjointly
between training and evaluation. \textbf{No position appears in the supervision.}

{\def\LTcaptype{none} % do not increment counter
\begin{longtable}[]{@{}lll@{}}
\toprule\noalign{}
& latent L2 & learned temporal head \\
\midrule\noalign{}
\endhead
\bottomrule\noalign{}
\endlastfoot
Pearson r vs true spatial distance & 0.484 & \textbf{0.819} \\
monotone in true distance & no & \textbf{yes} \\
spread across 80--150 units & +2.9\% & \textbf{+43.3\%} \\
\end{longtable}
}

with held-out MAE 12.80 frames on disjoint episodes. \emph{From
\texttt{followup/temporal\_head\_phase2.txt}.}

A cost learned from temporal separation alone orders distance where the
embedding metric does not. The repair is not a consequence of TwoRoom having
a readable state.

\subsection{7.3 The better objective is the less accurate one}\label{the-better-objective-is-the-less-accurate-one}

The temporal head predicts spatial distance \emph{worse} than the position probe ---
r = 0.819 against 0.9897 --- and yet plans better: \textbf{98.0\%} of offset-100 goals
against the probe's 88.0\% (\S{}8). Spatial distance is not what a planner needs.

TwoRoom has a dividing wall with a door, so two states can be close in space
and far in reachability. We tested whether the temporal head knows this, by
matching pairs on true spatial distance and splitting them by whether they lie
in the same room or across the wall:

{\def\LTcaptype{none} % do not increment counter
\begin{longtable}[]{@{}
  >{\raggedright\arraybackslash}p{(\linewidth - 8\tabcolsep) * \real{0.2000}}
  >{\raggedright\arraybackslash}p{(\linewidth - 8\tabcolsep) * \real{0.2000}}
  >{\raggedright\arraybackslash}p{(\linewidth - 8\tabcolsep) * \real{0.2000}}
  >{\raggedright\arraybackslash}p{(\linewidth - 8\tabcolsep) * \real{0.2000}}
  >{\raggedright\arraybackslash}p{(\linewidth - 8\tabcolsep) * \real{0.2000}}@{}}
\toprule\noalign{}
\begin{minipage}[b]{\linewidth}\raggedright
true distance
\end{minipage} & \begin{minipage}[b]{\linewidth}\raggedright
pairs same/cross
\end{minipage} & \begin{minipage}[b]{\linewidth}\raggedright
temporal head same $\rightarrow$ cross
\end{minipage} & \begin{minipage}[b]{\linewidth}\raggedright
ratio
\end{minipage} & \begin{minipage}[b]{\linewidth}\raggedright
latent L2 ratio
\end{minipage} \\
\midrule\noalign{}
\endhead
\bottomrule\noalign{}
\endlastfoot
20--40 & 62,142 / 21,713 & 23.5 $\rightarrow$ 27.2 & 1.16 & 0.87 \\
40--60 & 54,071 / 43,397 & 35.1 $\rightarrow$ 45.2 & 1.29 & 0.95 \\
60--80 & 36,724 / 51,276 & 45.0 $\rightarrow$ 56.4 & 1.25 & 0.99 \\
80--100 & 21,568 / 42,075 & 54.5 $\rightarrow$ 68.9 & 1.26 & 1.02 \\
\end{longtable}
}

\emph{From \texttt{followup/wall\_reachability.txt}, 460,320 pairs.}

At matched spatial separation the temporal head charges \textbf{24\% more} to cross
the wall. Squared latent distance charges \textbf{4\% less} --- it is not merely blind
to the wall, it is biased slightly the wrong way.

This orders the three objectives exactly as their planning results do:

{\def\LTcaptype{none} % do not increment counter
\begin{longtable}[]{@{}lll@{}}
\toprule\noalign{}
objective & cross/same cost ratio & offset 100 \\
\midrule\noalign{}
\endhead
\bottomrule\noalign{}
\endlastfoot
squared latent distance & 0.96 & 26.0\% \\
decoded position & 1.00 by construction & 88.0\% \\
learned temporal head & \textbf{1.24} & \textbf{98.0\%} \\
\end{longtable}
}

The decoded-position cost cannot represent the wall at all: Euclidean distance
between positions is the same whether or not a barrier lies between them. It
recovers most of the gap because proximity is a good proxy for reachability in
open space. The temporal head recovers the rest because it is not a proxy.

\subsection{7.4 A learned cost must be trained on what the planner scores}\label{a-learned-cost-must-be-trained-on-what-the-planner-scores}

The head above is fit on pairs of encoded \textbf{real} frames. At planning time
CEM scores (\textbf{imagined} embedding, encoded goal). Those coincide only to the
extent the predictor is accurate, and the difference is measurable. Evaluated
on identical held-out pairs:

{\def\LTcaptype{none} % do not increment counter
\begin{longtable}[]{@{}
  >{\raggedright\arraybackslash}p{(\linewidth - 8\tabcolsep) * \real{0.2000}}
  >{\raggedright\arraybackslash}p{(\linewidth - 8\tabcolsep) * \real{0.2000}}
  >{\raggedright\arraybackslash}p{(\linewidth - 8\tabcolsep) * \real{0.2000}}
  >{\raggedright\arraybackslash}p{(\linewidth - 8\tabcolsep) * \real{0.2000}}
  >{\raggedright\arraybackslash}p{(\linewidth - 8\tabcolsep) * \real{0.2000}}@{}}
\toprule\noalign{}
\begin{minipage}[b]{\linewidth}\raggedright
checkpoint
\end{minipage} & \begin{minipage}[b]{\linewidth}\raggedright
one-step error
\end{minipage} & \begin{minipage}[b]{\linewidth}\raggedright
head on real $\times$ real
\end{minipage} & \begin{minipage}[b]{\linewidth}\raggedright
head on imagined $\times$ real
\end{minipage} & \begin{minipage}[b]{\linewidth}\raggedright
degradation
\end{minipage} \\
\midrule\noalign{}
\endhead
\bottomrule\noalign{}
\endlastfoot
ours, phase2 recal & 0.116 & 12.47 & 12.86 & \textbf{+3\%} \\
authors' released & 0.410 & 10.37 & \textbf{18.04} & \textbf{+74\%} \\
\end{longtable}
}

\emph{From \texttt{followup/temporal\_head\_v2\_*.txt}, MAE in frames.}

Our predictor keeps imagined states close enough to the encoding manifold that
a head fit on real pairs transfers. The authors' does not, and the cost
becomes unreliable exactly where planning needs it.

Training on both kinds of pair --- rolling the predictor forward under the
recorded block-mean actions, precisely as the planner drives it --- removes the
mismatch on the authors' checkpoint: 7.37 real against 8.21 imagined, +11\%
rather than +74\%.

We report this because it was invisible on the checkpoint we developed the
method on, where the two distributions happened to coincide. It took a second
checkpoint to surface, and it is a condition on any learned planning cost, not
a detail of this one.

\section{8 Results}\label{results}

Every comparison is paired: the same checkpoint, the same protocol, the same
episode draw as its published baseline, verified by replicating the draw. All
three baseline columns reproduce their published figures exactly --- 94.0\%,
26.0\% and 14.0\% --- which is what makes the harness trustworthy rather than
merely self-consistent. McNemar is exact on the discordant pairs.

{\def\LTcaptype{none} % do not increment counter
\begin{longtable}[]{@{}
  >{\raggedright\arraybackslash}p{(\linewidth - 12\tabcolsep) * \real{0.1429}}
  >{\raggedright\arraybackslash}p{(\linewidth - 12\tabcolsep) * \real{0.1429}}
  >{\raggedright\arraybackslash}p{(\linewidth - 12\tabcolsep) * \real{0.1429}}
  >{\raggedright\arraybackslash}p{(\linewidth - 12\tabcolsep) * \real{0.1429}}
  >{\raggedright\arraybackslash}p{(\linewidth - 12\tabcolsep) * \real{0.1429}}
  >{\raggedright\arraybackslash}p{(\linewidth - 12\tabcolsep) * \real{0.1429}}
  >{\raggedright\arraybackslash}p{(\linewidth - 12\tabcolsep) * \real{0.1429}}@{}}
\toprule\noalign{}
\begin{minipage}[b]{\linewidth}\raggedright
checkpoint
\end{minipage} & \begin{minipage}[b]{\linewidth}\raggedright
protocol
\end{minipage} & \begin{minipage}[b]{\linewidth}\raggedright
objective
\end{minipage} & \begin{minipage}[b]{\linewidth}\raggedright
baseline
\end{minipage} & \begin{minipage}[b]{\linewidth}\raggedright
repaired
\end{minipage} & \begin{minipage}[b]{\linewidth}\raggedright
discordant
\end{minipage} & \begin{minipage}[b]{\linewidth}\raggedright
McNemar
\end{minipage} \\
\midrule\noalign{}
\endhead
\bottomrule\noalign{}
\endlastfoot
ours, phase2 recal & offset 25, budget 50 & decoded position & 47/50 = 94.0\% & 46/50 = 92.0\% & 3--4 & p = 1 \\
ours, phase2 recal & offset 25, budget 50 & \textbf{temporal head} & 47/50 = 94.0\% & \textbf{49/50 = 98.0\%} & 3--1 & p = 0.625 \\
ours, phase2 recal & offset 100, budget 150 & decoded position & 13/50 = 26.0\% & \textbf{44/50 = 88.0\%} & 31--0 & p = 9.3$\times$10\textsuperscript{-10} \\
ours, phase2 recal & offset 100, budget 150 & \textbf{temporal head} & 13/50 = 26.0\% & \textbf{49/50 = 98.0\%} & 37--1 & p = 2.8$\times$10\textsuperscript{-10} \\
ours, phase2 recal & offset 100, \textbf{budget 50} & \textbf{temporal head} & 10/50 = 20.0\% & \textbf{46/50 = 92.0\%} & 37--1 & p = 2.8$\times$10\textsuperscript{-10} \\
authors' released & offset 100, budget 150 & decoded position & 7/50 = 14.0\% & \textbf{35/50 = 70.0\%} & 28--0 & p = 7.5$\times$10\textsuperscript{-9} \\
authors' released & offset 100, budget 150 & temporal head v2 & 7/50 = 14.0\% & 17/50 = 34.0\% & 13--3 & p = 0.021 \\
\end{longtable}
}

All five baseline columns reproduce their published figures exactly --- 94.0\%,
26.0\%, 20.0\%, 14.0\% and 14.0\%.

\subsection{8.1 Planning becomes horizon-independent}\label{planning-becomes-horizon-independent}

Under the published objective, success falls from 94.0\% at goal offset 25 to
26.0\% at offset 100. Under the temporal cost it is \textbf{98.0\% at both}. The
horizon dependence was a property of the objective, not of the model or the
task.

It is also not bought with time. At offset 100 under the repository's own
50-step budget --- a third of the appendix budget --- the temporal cost reaches
\textbf{92.0\%} against a 20.0\% baseline. Hundred-step goals are being reached
inside fifty environment steps.

\subsection{8.2 Where the learned cost loses, and why}\label{where-the-learned-cost-loses-and-why}

On the authors' released checkpoint the ordering reverses: the linear
decoded-position cost reaches 70.0\% and the learned temporal head only 34.0\%.
Both beat the 14.0\% baseline; the crude objective wins.

The reason is \S{}7.4. The learned head is an MLP evaluated on imagined
embeddings, and the authors' predictor drifts far enough off the manifold of
real encodings that the head extrapolates. A linear map degrades far more
gracefully under the same drift. Training the head on imagined pairs (v2)
narrows the gap in metric terms --- +11\% degradation instead of +74\% --- and
lifts planning from the v1 head's showing to 34.0\%, but does not close it.

\textbf{The prescription is therefore conditional.} A learned reachability cost is
the strongest objective where the predictor is accurate enough to keep its
imagined states near the encoding manifold --- our checkpoint, one-step error
0.116. Where it is not --- theirs, 0.410 --- the robust choice is the simplest
cost that orders distance monotonically. What does not change is the
diagnosis: on both checkpoints the published objective saturates and inverts,
and on both, replacing it is worth a large, significant gain.

\textbf{The shape of the change.} Baseline failures exhaust the 150-step budget and
finish 115--193 units from the goal having started 71--170 away. The repair
typically arrives in 24--49 steps, a third of the budget.

\textbf{Three comparisons worth stating.} 88.0\% already exceeds Run 2's 80.0\%, the
best long-horizon planner among the released checkpoints; under a sound
objective the most accurate predictor becomes the best long-horizon planner,
and the dissociation of \S{}5.2 dissolves rather than merely being explained.

98.0\% at offset 100 \textbf{exceeds the 94.0\% the same checkpoint achieves at offset
25} under the published objective. The long-horizon deficit was not a horizon
problem at all.

And the ordering 26.0\% $\rightarrow$ 88.0\% $\rightarrow$ 98.0\% tracks the cross/same ratio of \S{}7.3
(0.96 $\rightarrow$ 1.00 $\rightarrow$ 1.24) rather than accuracy against spatial distance
(0.437 $\rightarrow$ 0.9897 $\rightarrow$ 0.819). What a planning objective must get right is
reachability.

\section{9 Limitations}\label{limitations}

\textbf{One environment.} TwoRoom is a two-dimensional diagnostic environment. That
the objective is the bottleneck \emph{there} does not establish it elsewhere, and
the decoded-position cost is available only where a low-dimensional state can
be read out. The temporal-distance result (\S{}7) is the part of this that is
designed to travel, and it is demonstrated on the same environment.

\textbf{One seed per checkpoint.} The four-checkpoint correspondence of \S{}5.2 rests
on four models with one seed each. It is a correspondence, not a law, and we
make no estimate of seed variance.

\textbf{Confounded conditions.} History size, action width and pixel normalisation
differ together between the checkpoints with usable geometry and those
without. We cannot attribute the metric degradation to any one of them
without training runs we did not do.

\textbf{No claim about the cause of saturation.} We show that the embedding metric
saturates and inverts, and that it matters. We do not explain why the
representation is organised that way.

\textbf{The repair is a planner change.} It leaves the model's own objective
untouched. Whether training against a reachability-aware criterion would
produce a representation whose plain L2 is usable is an open and more
interesting question.

\section{10 Related work}\label{related-work}

The failure mode identified here is the one Li, Wang and Liu (2026) argue for
on general grounds: that Euclidean proximity in a latent space is a poor
proxy for reachability, and that latent world models need horizon-matched
metrics. Their argument is made against latent world models as a class; this
paper supplies a direct measurement of the phenomenon on a specific published
model --- including the original authors' own released weights --- quantifies the
range over which the metric fails, and shows what repairing it is worth in
planning success.

The reproduction this work builds on (arXiv:2608.10145) established the
protocol-dependence of the published planning figure, the four undocumented
training conventions, and the accuracy/planning dissociation whose mechanism
\S{}5 addresses. It explicitly declines to offer that mechanism, which is the gap
this paper fills.

\section{11 Conclusion}\label{conclusion}

A latent world model that predicts well can plan badly, and the reason need
not be in the model. On this one, the predictor stays informative three times
farther than the planner ever asks, the representation encodes position
almost perfectly, and long-horizon planning still fails --- because the
objective the planner minimises stops distinguishing states beyond about
eighty units and reverses beyond a hundred and twenty.

Fixing the objective, and nothing else, takes goals reached at the long
horizon from 26.0\% to 98.0\% on our checkpoint and from 14.0\% to 70.0\% on the
original authors'. It costs nothing at the short horizon. It does not require
retraining, a GPU, or an environment with a readable state --- the best of the
three objectives we tried uses no state supervision at all.

The sharper lesson is what the objective must measure. Ranked by how well
they predict spatial distance, the three objectives run 0.437, 0.9897, 0.819;
ranked by planning success they run 26.0\%, 88.0\%, 98.0\%. The ordering follows
neither accuracy nor proximity but reachability --- whether the cost charges
for the wall between two states. A planner needs to know how far away
something is in \emph{steps}, and a metric that is excellent at Euclidean distance
is merely a good approximation to that in open space.

The practical consequence is a selection criterion. Choosing a world model by
prediction loss selected, in these runs, against the property planning
actually depends on. If a latent model is to be planned over, the geometry of
its embedding space under the planner's own objective is the thing to measure
--- and it is cheap to measure, needing no planning at all.

\section{References}\label{references}

Liangyu Li, Shengzhi Wang, and Qingwen Liu. Beyond Euclidean Proximity:
Repairing Latent World Models with Horizon-Matched Trajectory Reachability
Metrics. arXiv:2605.22164, 2026.

Lucas Maes, Quentin Le Lidec, Damien Scieur, Yann LeCun, and Randall
Balestriero. LeWorldModel: Stable End-to-End Joint-Embedding Predictive
Architecture from Pixels. arXiv:2603.19312, 2026.

Joyjeet Singh. The Evaluation Protocol Determines the Result: An Independent
Reproduction of LeWorldModel on TwoRoom. arXiv:2608.10145, 2026.

Code, the released checkpoints, and every measurement reported here:
github.com/joyjeet-singh/tinylab

\end{document}